\documentclass[conference]{IEEEtran}
\usepackage{blindtext}
\usepackage[a4paper, includeheadfoot, margin=1.5cm]{geometry}
\usepackage[utf8]{inputenc}
\usepackage[english]{babel}
\usepackage{textcomp}

\usepackage{ltxgrid}
\usepackage{float}
\usepackage{graphicx}
\usepackage{subcaption}
\usepackage{caption}
\usepackage[T1]{fontenc}
\usepackage{fancyhdr}
\usepackage[cmex10]{amsmath}
\usepackage{mdwmath}
\usepackage{eqparbox}
\usepackage{gensymb}
\usepackage{amssymb}

\usepackage[style=numeric,sorting=none,backend=biber]{biblatex}
\usepackage[hidelinks]{hyperref}
\begin{document}

\title{ {\fontfamily{bch}\selectfont \textbf{Studying Invariances of Trained \\ Convolutional Neural Networks}}\vspace{-1em}}
\author{
    \IEEEauthorblockN{Charlotte Bunne\IEEEauthorrefmark{1}\IEEEauthorrefmark{3}, Lukas Rahmann\IEEEauthorrefmark{1}\IEEEauthorrefmark{3}, and Thomas Wolf\IEEEauthorrefmark{1}\IEEEauthorrefmark{3}}
    \IEEEauthorblockA{\IEEEauthorrefmark{3}Eidgenössische Technische Hochschule (ETH) Zürich \\ \\
    \{bunnec, lrahmann, wolftho\}@ethz.ch} \\
    \IEEEauthorblockA{\IEEEauthorrefmark{1}These authors contributed equally to this work.} \\
}
\maketitle

\twocolumngrid
\begin{abstract}
\boldmath
Convolutional Neural Networks (CNNs) define an exceptionally powerful class of models for image classification, but the theoretical background and the understanding of how invariances to certain transformations are learned is limited.
In a large scale screening with images modified by  different affine and nonaffine transformations of varying magnitude, we analyzed the behavior of the CNN architectures \textit{AlexNet} and \textit{ResNet}. If the magnitude of different transformations does not exceed a class- and transformation dependent threshold, both architectures show invariant behavior.
In this work we furthermore introduce a new learnable module, the \textit{Invariant Transformer Net}, which enables us to learn differentiable parameters for a set of  affine transformations. This allows us to extract the space of transformations to which the CNN is invariant and its class prediction robust.
\end{abstract}


\IEEEpeerreviewmaketitle
\pagestyle{fancy}

\section{Introduction}
Convolutional Neural Networks have demonstrated impressive performance in the field of computer vision and natural language processing and are thus the cutting edge method for image classification. In several layers of trainable convolutions and subsampling interspersed with sigmoid nonlinearity, features of the input are extracted and fed into a trainable classifier.
To classify images or solve pattern recognition tasks reliably, a CNN should combine invariance and discriminability to the input variable. While high level features of the input should be learned, the network's prediction should be robust to irrelevant input transformations. Thus, learning selective invariant features is a difficult task \cite{goodfellow2009}. Despite the inflationary use of CNNs, the mathematical theory of how features are extracted and invariances to certain transformations are learned is not well understood yet \cite{wiatowski2017}.

In this paper, we propose a new method to extract the space of possible transformations $F$  such that for any image $I$, the transformed image $F(I)$ and $I$ are classified similarly by the trained network. To achieve this, we propose the architecture \textit{Invariant Transform Net}, which allows us to introduce affine transformations specified by differentiable parameters and thus, access the space of possible image modifications to which the network is invariant.

We further evaluate the behavior of different CNN architectures by passing a set of affine and nonaffine transformations of increasing magnitude to the trained networks and analyze which kinds of invariances are present. This enables us to define thresholds of different transformations, which -- when exceeded -- lead to a change in the classification result.

\section{Related Work}

One of the earliest treatments of invariances in deep neural networks was concerned with the question of how to determine the quality of learned representations in an unsupervised fashion \cite{goodfellow2009}.
The authors argue that good representations should not only be able to achieve a high performance in a supervised setting (\emph{discriminability}), but also generally be invariant to certain transformations in the inputs.
They probe this invariance by defining an activity threshold for every hidden neuron based on its responses to random inputs and then applying transformations to inputs for which this neuron is considered active based on the threshold.
If the neuron stays active under the input transformations, it is called invariant to them.
They use translation and rotation (in 2D and 3D) from natural videos as transformations and test their method on stacked autoencoders and deep belief networks.
They observe that invariance to those transformations increases with the depth of the model architecture.

Another work studies invariance in learned representations as a special case of equivariance in general features \cite{lenc2015}.
The authors examine equivariance in a representation by trying to learn a mapping from input space transformations to transformations in the feature space.
In parts of the input space where this mapping approaches the identity function, the features can be considered invariant to the input transformations.
They use rotations, rescaling and flips of images and find that the equivariance of latent features in deep convolutional networks to those transformations decreases with depth, while the invariance interestingly reaches its maximum in the middle layers of the networks.
They also find that the representations in the first layers of different networks are largely equivalent to each other which is not the case for the deeper layers.

\section{ Models and Methods}
\subsection{Affine and Nonaffine Transformations}
Affine transformations map an input $\mathbf{x}$ from a space  $X$ into a space $Y$ using an affine map $F\colon x \to Y$. The affine transformation is of the form $x \mapsto Mx + b$, where $M$ is a linear transformation on $X$ and $b$ a vector in $Y$. Affine transformations include rotations, translations, as well as scaling. In affine transformations, parallel lines remain parallel \cite{szeliski2010}. Nonaffine transformations comprise Gaussian noise, Gaussian blur, whiteness shifts, contrast and other nonlinearities. In this paper, the behavior of a neural network to both, affine and nonaffine transformations is analyzed.

\subsection{Convolutional Neural Network Architectures}
Convolutional Neural Networks are deep, feed-forward artificial neural networks consisting of convolutional hidden, pooling, fully connected and normalization layers. Inspired by biological receptive fields of the visual cortex of animals, CNNs outperform traditional, hand-crafted feature approaches in tasks such as image and video recognition, recommender systems and natural language processing \cite{lecun1998, krizhevsky2012}.
Due to its simplicity and well studied behavior we chose the pretrained network from \cite{guerzhoy2016} with an \textit{AlexNet} architecture \cite{krizhevsky2012}. Furthermore, we used a pretrained network following the \textit{ResNet} architecture (\texttt{ResNet V1 101}) published in the \textit{TensorFlow-Slim image classification model library}. The \textit{ResNet} architecture consists of 152 layers and is winner of the  \textit{ImageNet Large Scale Visual Recognition Challenge} (ILSVRC) 2015 \cite{he2016}.

To analyze the behavior of the CNNs \textit{AlexNet} and \textit{ResNet}, we used the dataset published in ILSVRC, a well-known benchmark dataset in object category classification consisting of millions of images for the detection of hundreds of object categories \cite{russakovsky2015}.

\subsection{Strategy 1: Sensibility To Transformations}
Using the pretrained \textit{AlexNet}, we apply affine and nonaffine transformations $F(I,v): [0,255]^{W \times H \times 3} \to [0,255]^{W \times H \times 3}$ of increasing magnitude $v$ on the input images $I(x,y)$ to get $I'(x,y)=F(I(x,y),v)$. Let $I \in [0,255]^{W \times H \times 3}$. We are considering the following transformations:
\begin{enumerate}
\item \textbf{Translation.} We translate the image either horizontally or vertically with $F_{trans_x}(I,v)(x,y) = I(x-vW,y)$ and $F_{trans_y}(I,v)(x,y) = I(x,y-vH)$
\item \textbf{Rotation.} To rotate the image, we apply $F_{rot}(I,v)(x,y) = I(rot((x,y),-v))$, where $rot((x,y),\phi)$ rotates the coordinates $(x,y)$ the amount of $\phi$ degrees around the center of the image.
\item \textbf{Scale.} We scale either in $x$ or $y$ direction with $F_{scale_x}(I,v)(x,y) = I(vx-(v-1)W/2,y)$ and $F_{scale_y}(I,v)(x,y) = I(x,vy-(v-1)H/2)$ which scales the coordinates with the factor $v$ and keeps the image in the center.
\item \textbf{Zoom.} Here we zoom in the center of the image. This can be written as a composition of the scale in $x$ and $y$ direction. $F_{zoom}(I,v) = F_{scale_y}(F_{scale_x}(I,v),v)$
\item \textbf{Brightness.} We apply an additive bias to each color channel: $F_{brightness}(I,v) = I+255v$.
\item \textbf{Contrast.} To adjust contrast, we just multiply all color values: $v$: $F_{contrast}(I,v) = vI$.
\item \textbf{Grayscale.} We linear transition between the color image and the grayscale image $I_{gray}$, where $I_{gray}$ denotes the grayscale image of $I$ in each channel: $F_{gray}(I,v) = (1-v)I + I_{gray}$.
\item \textbf{Gaussian Blur.} We 2D-convolve the image with an Gaussian kernel $K_{\sigma}$ with a standard deviation of $\sigma$: $F_{blur}(I,v) = I*K_{v}$
\item \textbf{Gaussian Noise.} We add a noise value to each pixel: $F_{noise}(I,v)(x,y) = I(x,y) + 255z_{v}$ where $z_{\sigma} \sim G(0,\sigma)$ and $G(0,\sigma)$ denotes the normal distribution with mean $0$ and a standard deviation of $\sigma$.
\end{enumerate}

This enabled us to define thresholds and identify critical points at which the network is not able to output correct prediction results any more. We use two measures to quantify this:
\begin{enumerate}
\item First we evaluated the average accuracy over the predictions of a set of test images with various classes. This can be used to analyze the sensibility of the network to the applied transformations.
\item Furthermore, we inspect different sets of test images of the same class. Here we analyze the softmax output of the network depending on the applied transformations.
\end{enumerate}

\subsection{Strategy 2: Invariant Transformer Net}
Instead of simply trying out to  which transformations the network is invariant to, it would be even more interesting whether the network itself is capable of learning a family of transformations $F(k_1,k_2)$ where $k_i \in [0,1]^2$ for $i \in \{1,2\}$  which would create different transformation functions depending on the inputs, to which the network is invariant to.

\begin{figure}
\centering
\includegraphics[width=\columnwidth]{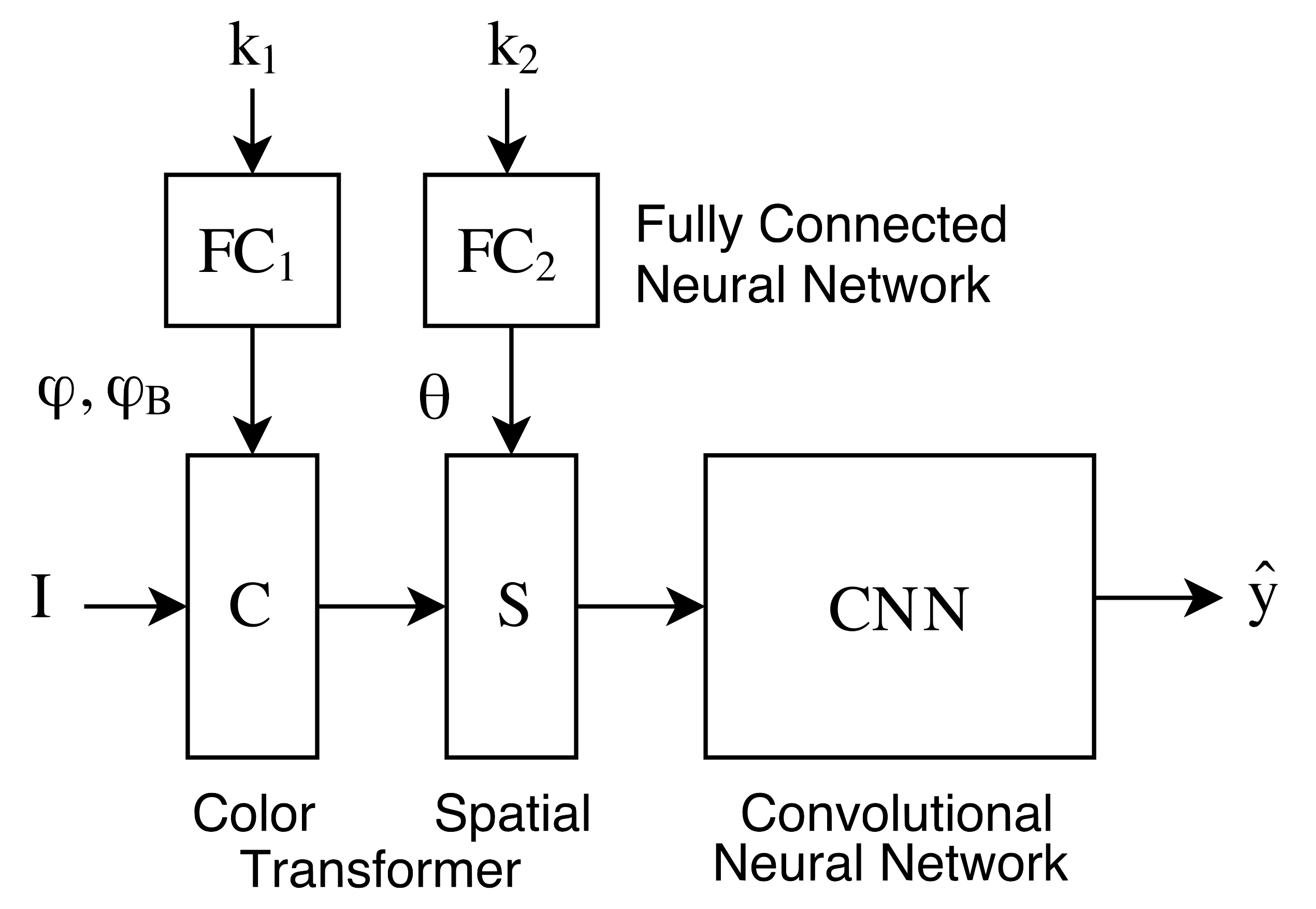}
\caption{\textbf{Architecture of the \textit{Invariant Transform Net}.} An input image $I$ is first transformed by an affine color transformation and afterwards by an affine spatial transformation. The resulting image is fed into a CNN network. The color and the spatial transformation are controlled parameters $\varphi$, $\varphi_B$ and $\theta$ which are generated by two dense layers controlled by two (two-dimensional) parameters $k_1 \in [0,1]^2$ and $k_2 \in [0,1]^2$.}
\label{fig:architecure}
\end{figure}

The network architecture we used is described in figure \ref{fig:architecure}. The overall idea is to train the weights of the layers $FC_1$ and $FC_2$ while keeping the weights of the CNN fixed. Each of those layers depends on a control parameter ($k_1$ and $k_2$). The parameters $k_1$ and $k_2$ are sampled during training uniformly at random from $[0,1]^2$ which allows us to learn a transformation function $F(k_1,k_2)$.

We decided to model color and spatial transformations. Both are differentiable and simple enough to avoid overfitting to single features of the images.
The layers $FC_1$ and $FC_2$ consist of two fully connected layers with rectified linear activation functions. This allows us to learn almost arbitrary functions which only depend on the parameters $k_i$. We decided to split the parameters $k_i$ between the spatial and color transformation such that both can be controlled independently.

While the color transformer is a simple matrix multiplication (in homogeneous coordinates to also allow brightness shifts) to the color values of the input image, the spatial transformer was taken from \cite{jaderberg2015} and enabled us to do differentiable (affine) transformations on the coordinates of the input images.

As both spatial and color transformations are affine, they can be described by two matrices $A_\theta\in \mathbb{R}^{2\times3}$ and $A_{\phi_B, \phi}\in \mathbb{R}^{3\times4}$.
We can further extend these to be quadratic, by setting the last row to $[0,0,...,1]$. We denote the extended matrices by $\hat{A}_\theta$ and $\hat{A}_{\phi_B, \phi}$

We describe a stochastic loss for these matrices, by measuring how far away they project a set of random unit vectors $\mathcal{S}$:
$$\hat{\mathcal{L}}(A) = - \frac{1}{\lvert S \rvert}\sum\limits_{x \in S} \lVert Ax - x\rVert_2^2$$
The more often and the further the matrix $A$ projects one of the elements in $x\in S$ away from its original position, the more this loss decreases.
This models the behavior we aim to achieve: to learn transformations of high magnitude such that the resulting transformed images are still classified correctly by the subsequent CNN.
In principle, other matrix norms can be used, but we achieved good results with this loss formulation.

Moreover, we also want to enforce that different $k_1$ and $k_2$ values lead to different generated functions by $FC_1$ and $FC_2$. We enforce this by incorporating the parameters $k_i$ to the loss function:

$$\mathcal{L}_{k_i}(A) = \sum\limits_{j \in \{0,1\}} k_{i,j} \hat{\mathcal{L}}(A)$$

where $k_{i,j}$ describes the $j$-th value of $k_i$.

The loss must further integrate the fact that learned transformations do not impact the prediction capabilities of the original network.
It proved to be rather difficult to train both at once, because of the extremely different value ranges these losses have. Thus, we decided to train the network using  batch wise accuracy to select whether we should increase the transformation impact or reduce the original loss.
So, for each batch running trough the network we select the batch wide loss to be:
$$\mathcal{L}_\text{final} = \begin{cases} \mathcal{L}_{
\textbf{orig}}   & acc < acc_{orig} \cr
c_\theta \mathcal{L}_{k_1}(\hat{A}_{\phi_B, \phi}) +  \mathcal{L}_{k_2}(\hat{A}_\theta)  & otherwise,
\end{cases}$$
where $acc_{orig}$ (accuracy) is selected based on the original performance of the used dataset and $c_\theta$ is an additional hyperparameter selected by hand to increase the influence of color transformations. This was needed because spatial transformations were learned much faster than color transformations.

While all tests were performed using the \textit{AlexNet} architecture, the \textit{Invariant Transformer Network} can be used with other CNN architectures as well.
The \textit{Invariant Transformer Network} is implemented in \textit{Tensorflow} (version 1.5.0) \cite{abadi2016}.

\section{Results}
\subsection{Strategy 1: Sensibility To Transformations}
The first approach analyzed the behavior of the networks \textit{AlexNet} as well as \textit{ResNet} for various affine and nonaffine transformations of differing magnitude.

For each plot we are using all images of a specific class of the ImageNet \cite{krizhevsky2012} test set and record the mean accuracy as well as the mean softmax output over all the images in one class while varying the parameters of the transformations. In each plot we show three predicted classes with the highest softmax output over all predicted classes while varying the parameter. If there are too many classes, we condense them to one line ("others") which shows the maximum softmax over all predicted classes for the given parameter.

\begin{figure*}[t]
\centering
\begin{subfigure}[t]{0.23\textwidth}
\includegraphics[width=\textwidth]{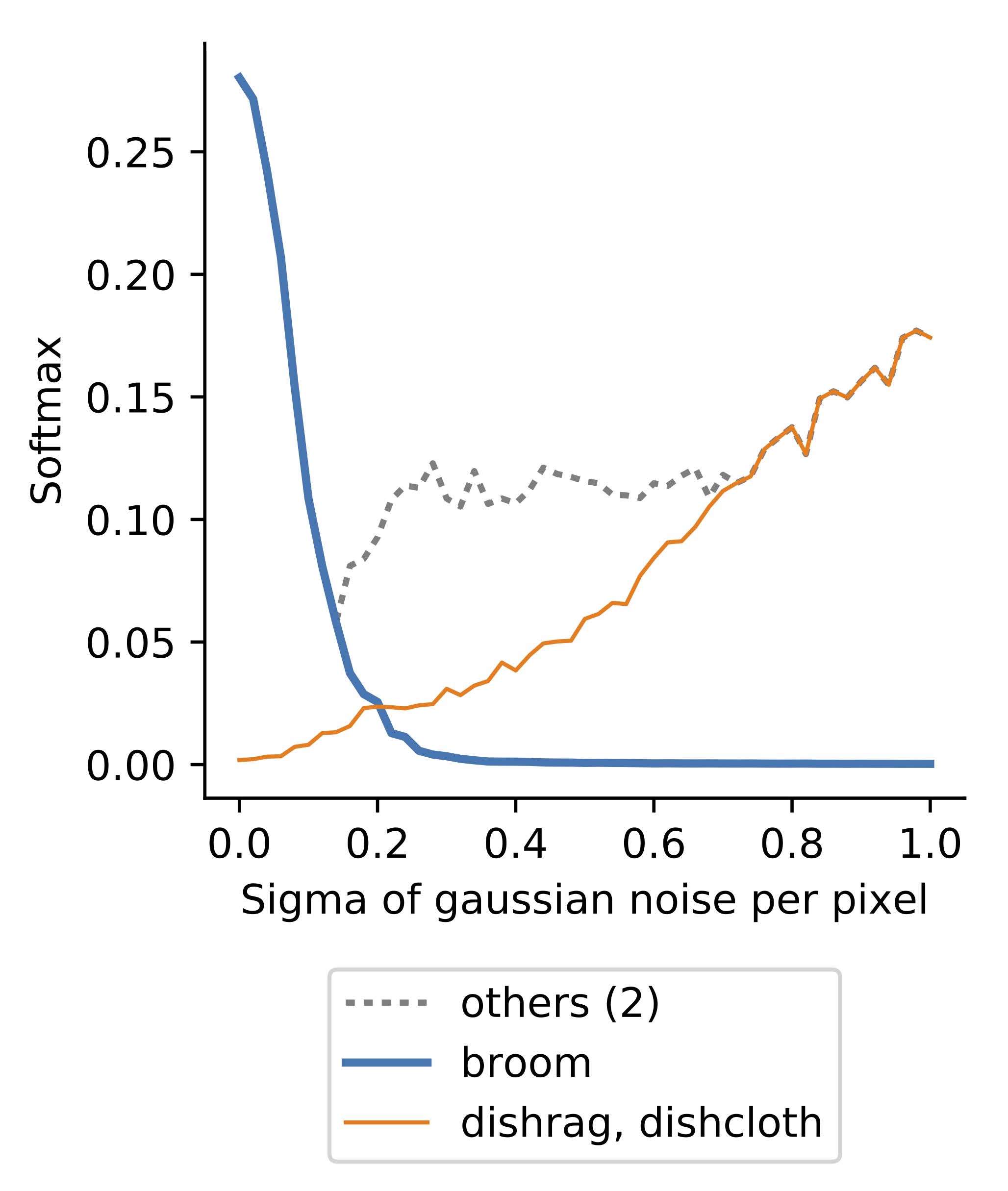}
\caption{Effect of increasing addition of Gaussian noise on the \textit{broom} class.}
\label{fig:plots_alex_gaussian}
\end{subfigure}%
\begin{subfigure}[t]{0.23\textwidth}
\includegraphics[width=\textwidth]{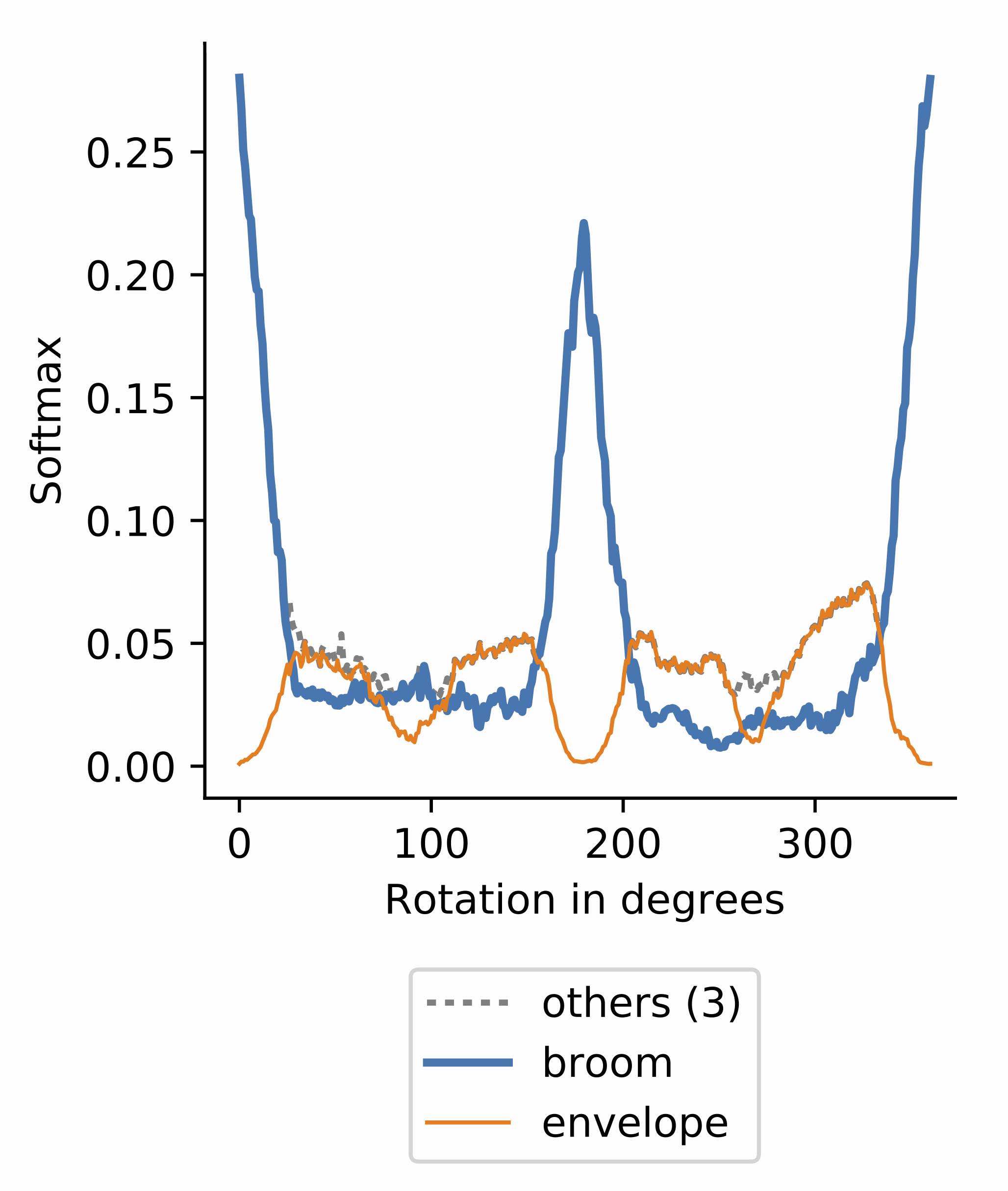}
\caption{Effect of rotation on the \textit{broom} class.}
\label{fig:plots_alex_rotation}
\end{subfigure}%
\begin{subfigure}[t]{0.23\textwidth}
\includegraphics[width=\textwidth]{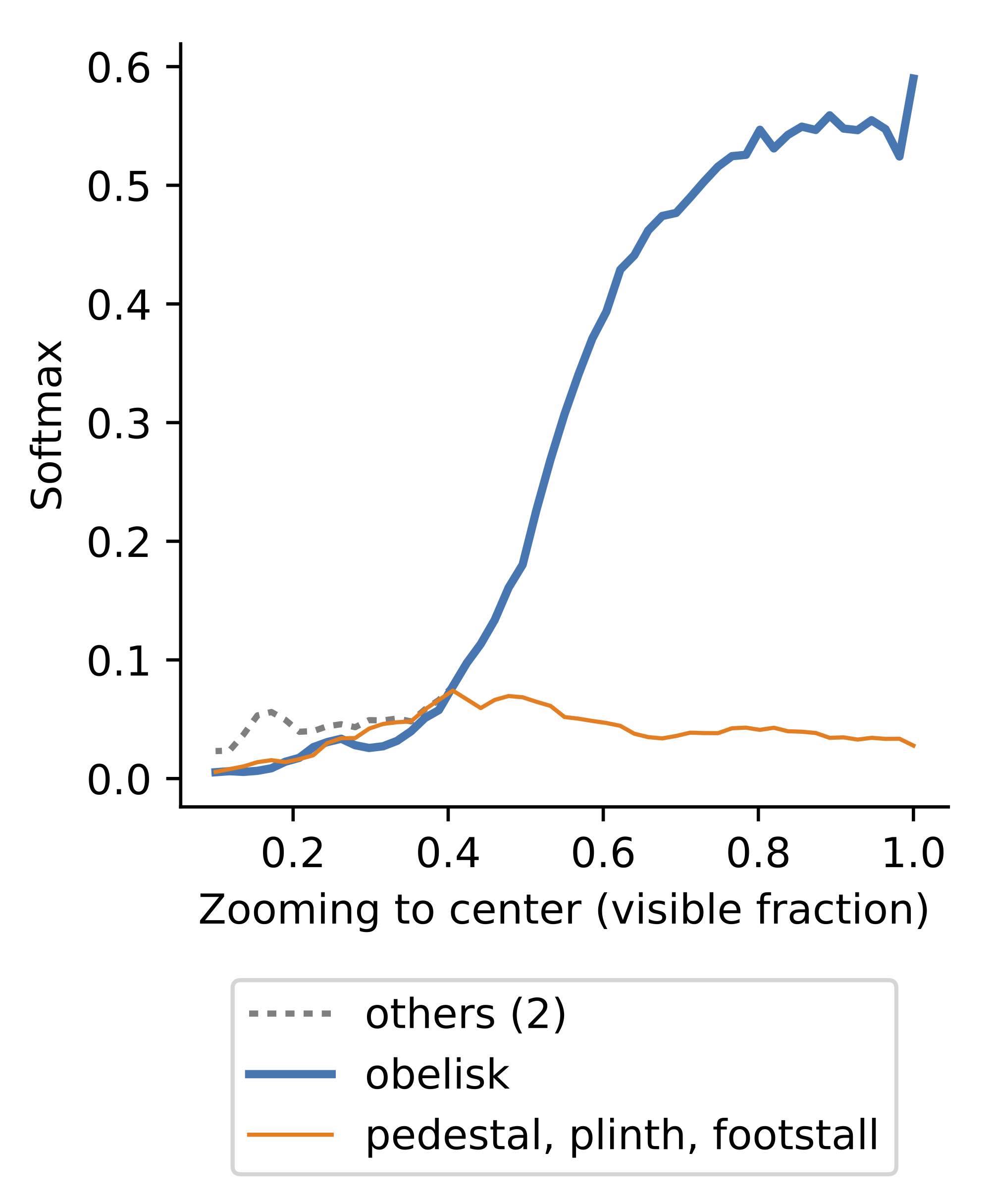}
\caption{Effect of zoom on the \textit{obelisk} class.}
\label{fig:}
\end{subfigure}%
\begin{subfigure}[t]{0.23\textwidth}
\includegraphics[width=\textwidth]{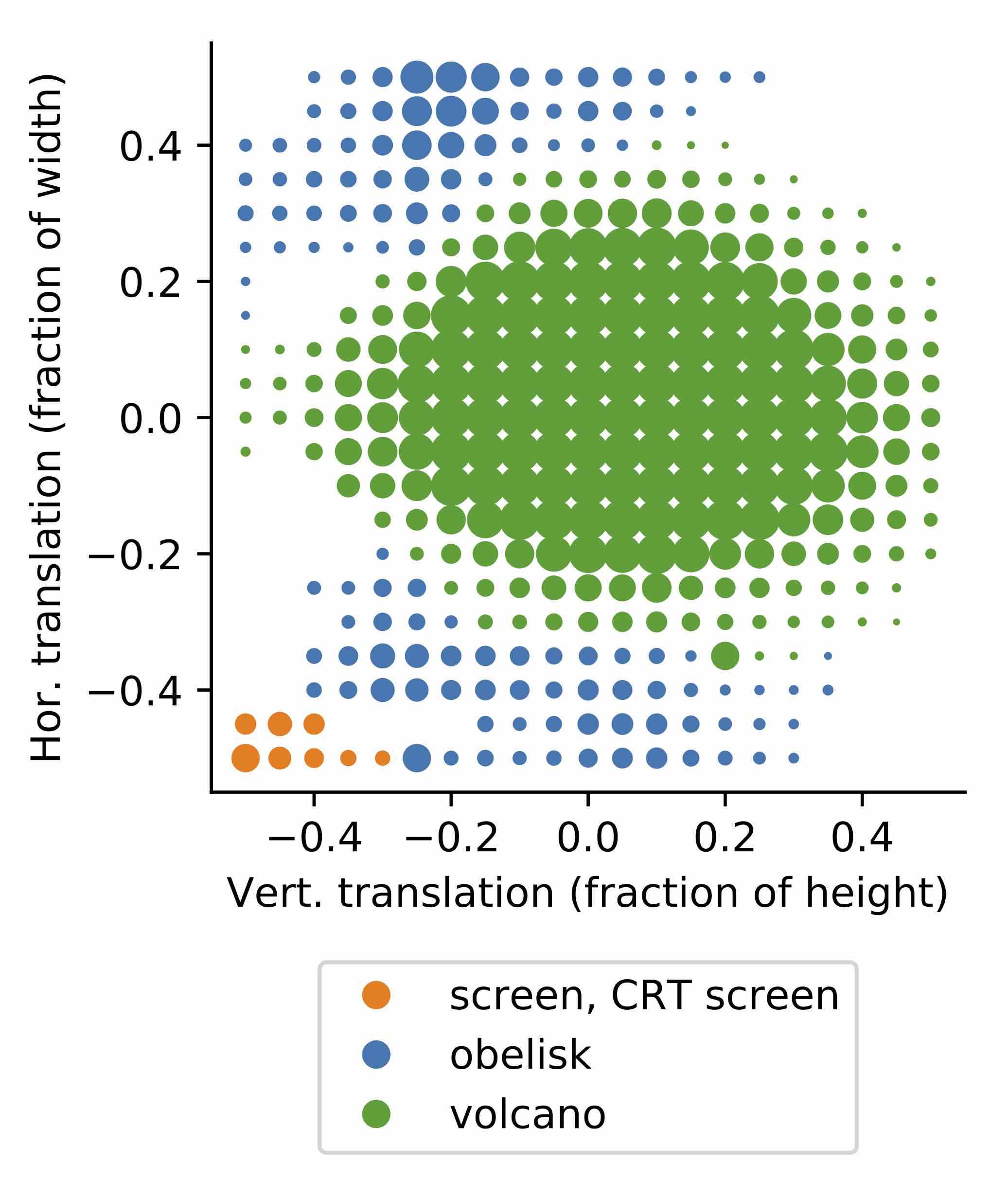}
\caption{Combined effect of horizontal and vertical translation on the \textit{volcano} class.}
\label{fig:}
\end{subfigure}
\caption{\textbf{Effect of affine and nonaffine transformations on classification results achieved by the \textit{AlexNet} architecture.} Class-wise average softmax outputs of the CNN with respect to increasing effects of different transformations. Here the effect of Gaussian noise, rotation  zooming and translation is visible.}
\label{fig:plots_alex}
\end{figure*}

Figure \ref{fig:plots_alex} shows exemplary results of Gaussian noise, rotation, zoom and translation on the performance of \textit{AlexNet}, figure \ref{fig:plots_res} on \textit{ResNet}. More effects of different transformations can be found in the appendix. Both network architecture show similar behavior with respect to the given transformation.

The results for both networks, \textit{AlexNet} and \textit{ResNet} suggest that the increasing addition of Gaussian noise leads to a switch of the class prediction as indicated by the softmax output. Since the networks' confidence is dependent on the class, the critical level of Gaussian noise leading to a switch in the prediction is dependent on the class as well (fig. \ref{fig:plots_alex_gaussian}). Rotation results in a strong reduction of the softmax. For certain point symmetric items like an orange, the softmax output is constant and independent on the transformation or for axisymmetric items, the softmax value returns to high values at a rotation of 180$\degree$ (fig. \ref{fig:plots_alex_rotation}).

\begin{figure*}[t]
\centering
\begin{subfigure}[t]{0.23\textwidth}
\includegraphics[width=\textwidth]{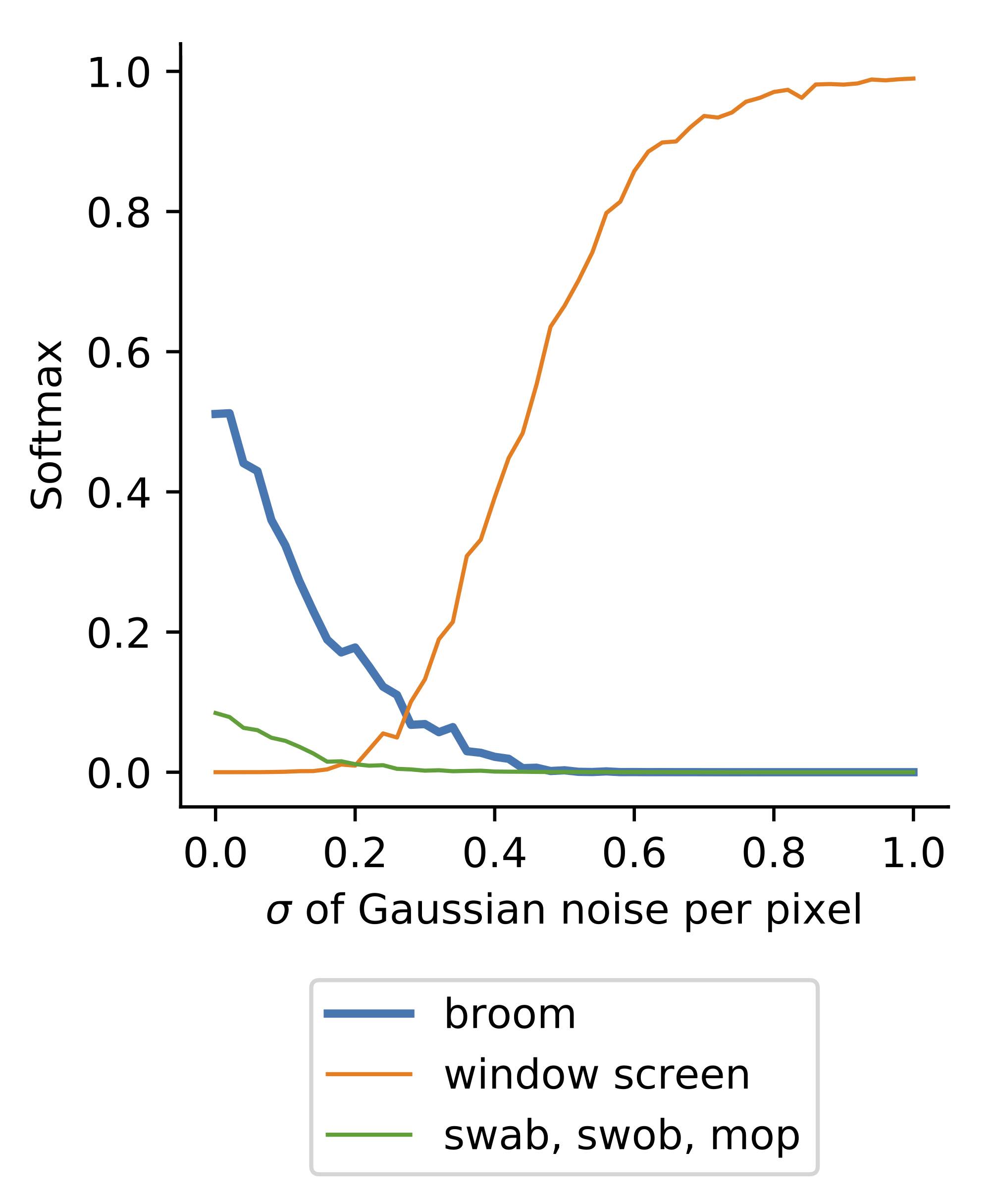}
\caption{Effect of increasing addition of Gaussian noise on the \textit{broom} class.}
\label{fig:}
\end{subfigure}%
\begin{subfigure}[t]{0.23\textwidth}
\includegraphics[width=\textwidth]{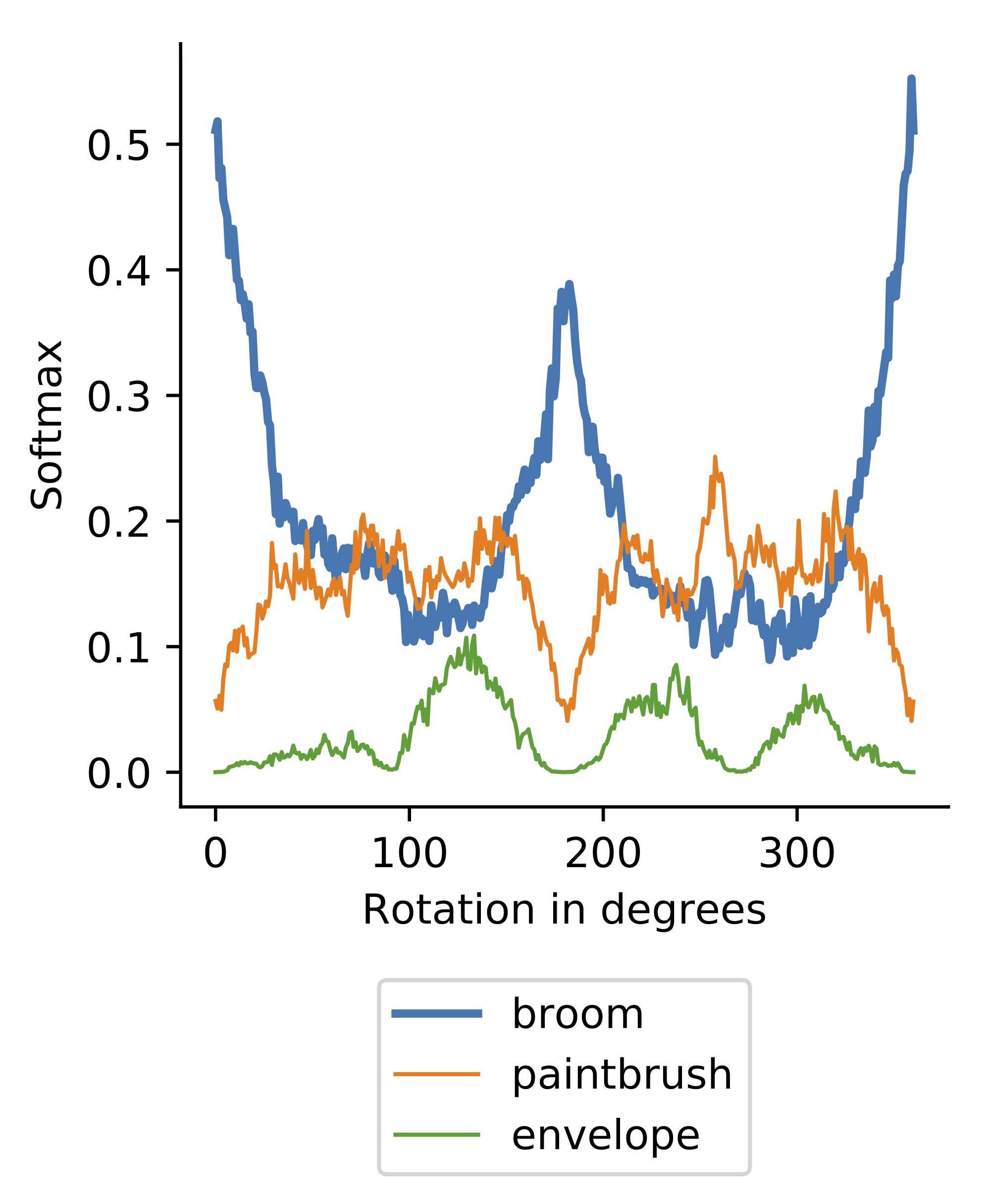}
\caption{Effect of rotation on the \textit{broom} class.}
\label{fig:tiger}
\end{subfigure}%
\begin{subfigure}[t]{0.23\textwidth}
\includegraphics[width=\textwidth]{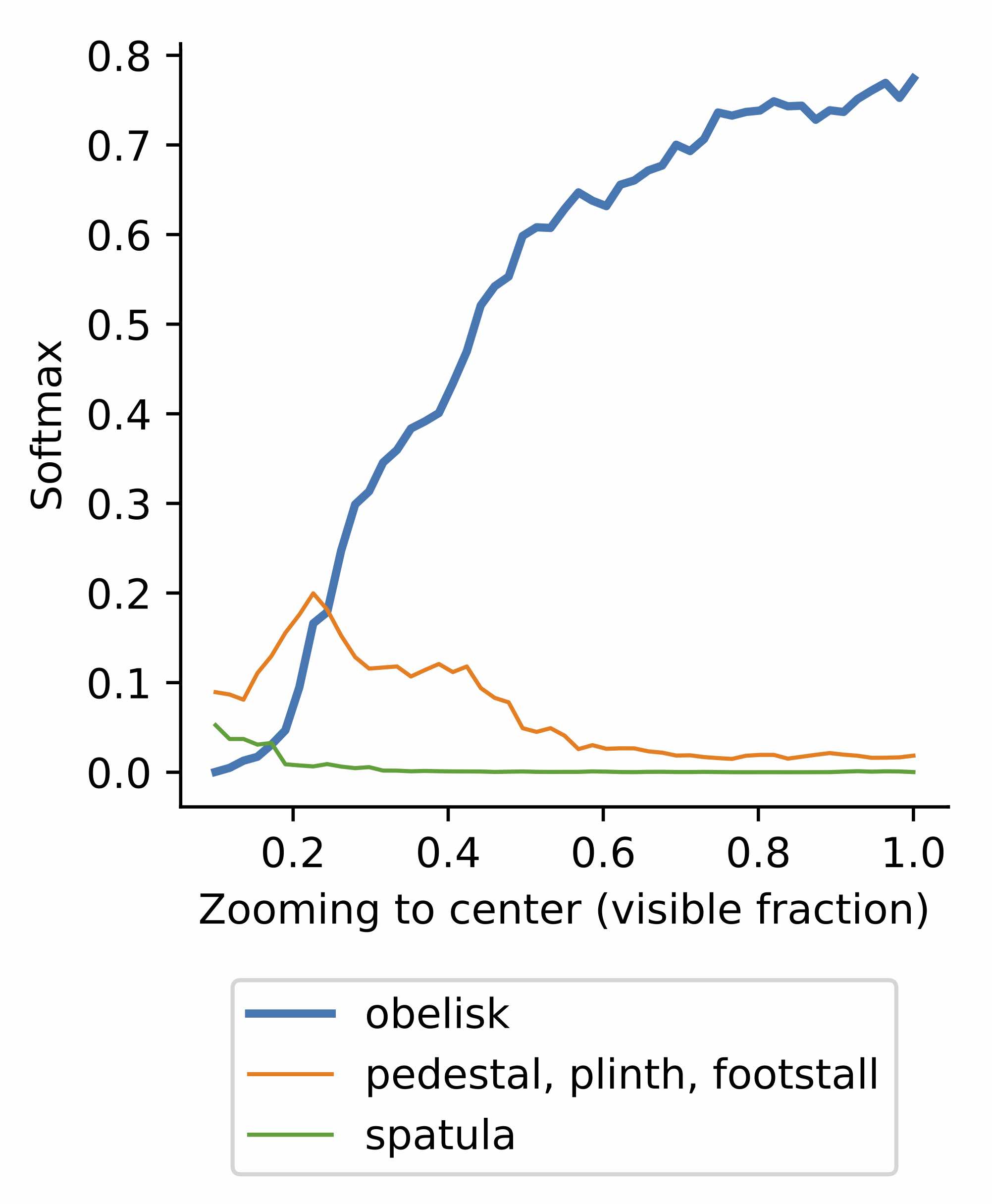}
\caption{Effect of zoom on the \textit{obelisk} class.}
\label{fig:}
\end{subfigure}%
\begin{subfigure}[t]{0.23\textwidth}
\includegraphics[width=\textwidth]{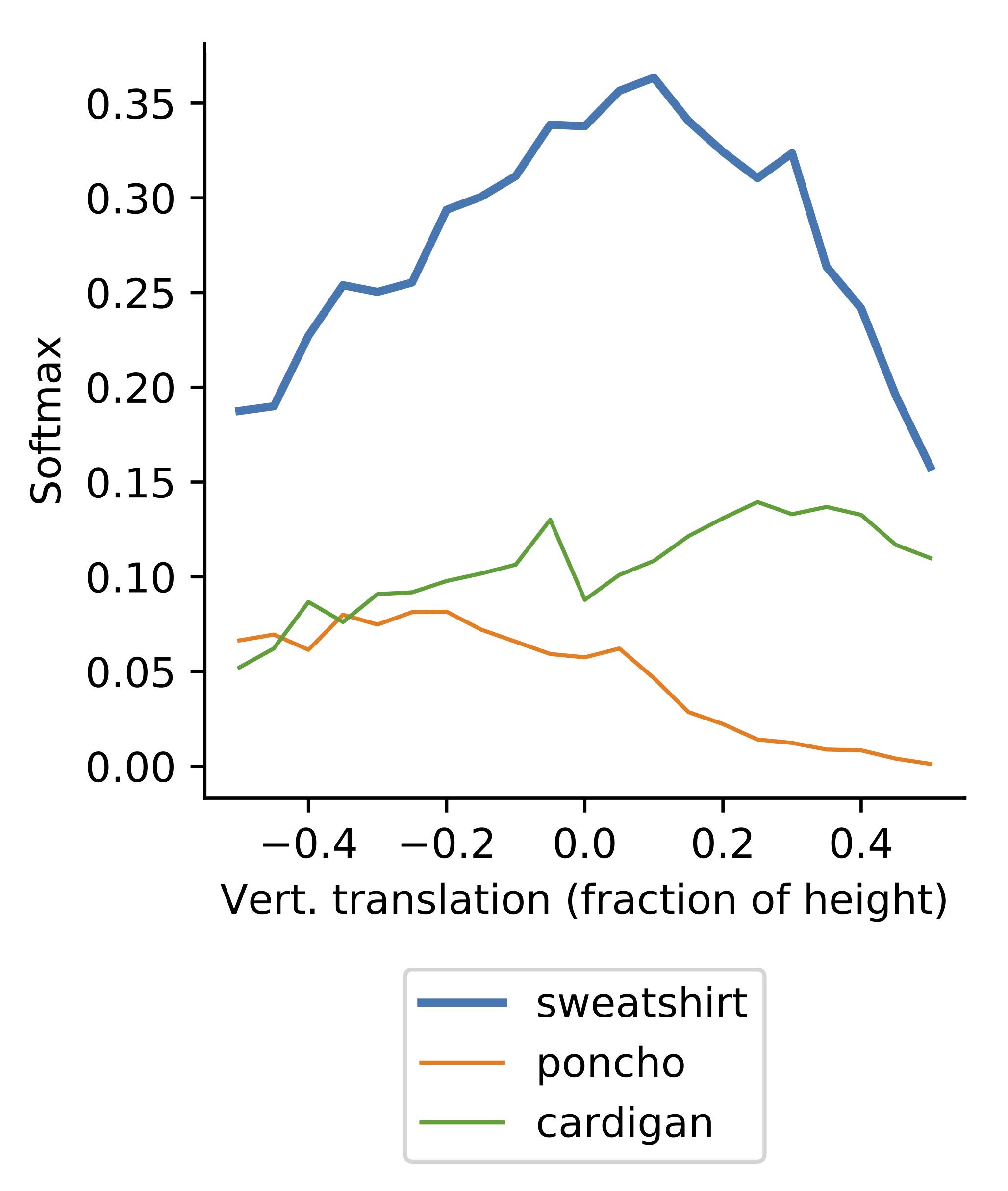}
\caption{Combinatorial effect of vertical translation on the \textit{sweatshirt} class.}
\label{fig:}
\end{subfigure}
\caption{\textbf{Effect of affine and nonaffine transformations on classification results achieved by the \textit{ResNet} architecture.} Class-wise average softmax outputs of the CNN with respect to increasing effects of different transformations. Here the effect of Gaussian noise, rotation  zooming and translation is visible.}
\label{fig:plots_res}
\end{figure*}

\subsection{Strategy 2: Invariant Transformer Net}
\begin{figure*}
\centering
\begin{subfigure}[t]{0.75\textwidth}
\includegraphics[width=0.9\textwidth]{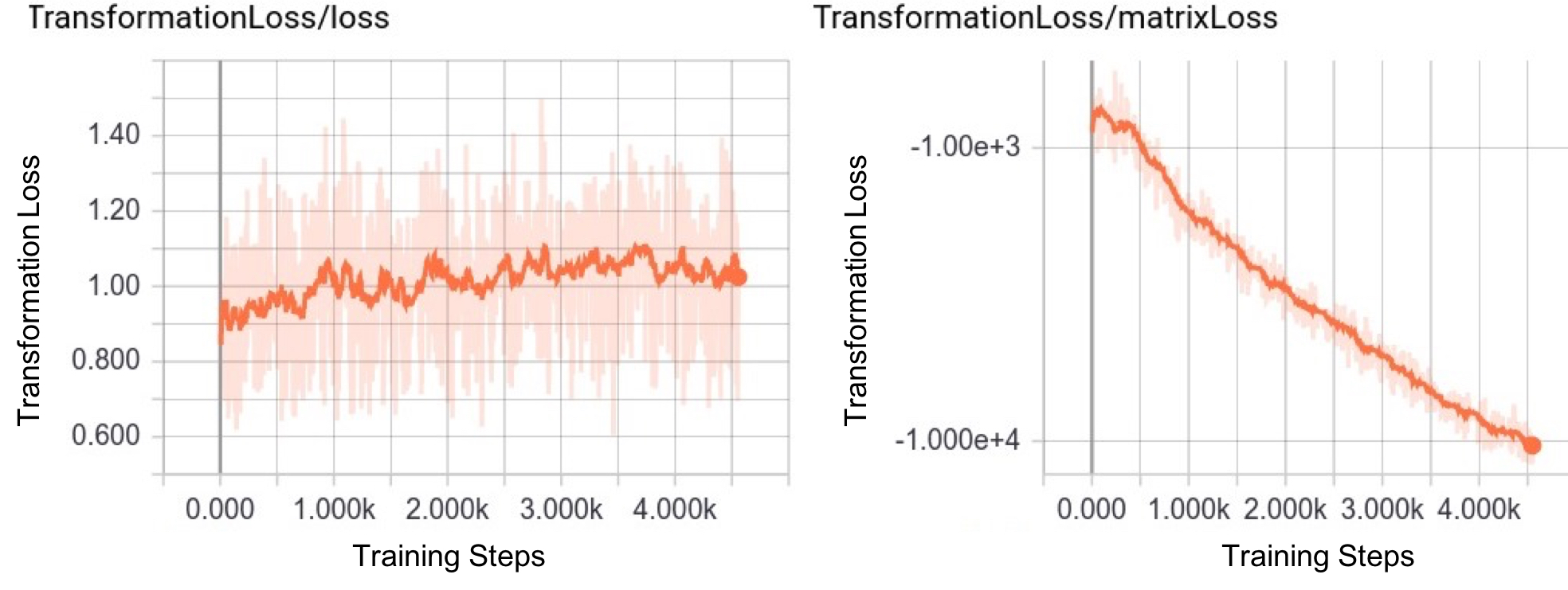}
\caption{Training the \textit{Invariant Transformer Net}: \\ The image shows the loss $\mathcal{L}_{\textbf{orig}}$ of the CNN (left) and $c_\theta \mathcal{L}_{k_1}(\varphi;\varphi_B) +  \mathcal{L}_{k_2}(\theta)$ (right).}
\label{fig:training}
\end{subfigure}%
\begin{subfigure}[t]{0.23\textwidth}
\includegraphics[width=0.9\textwidth]{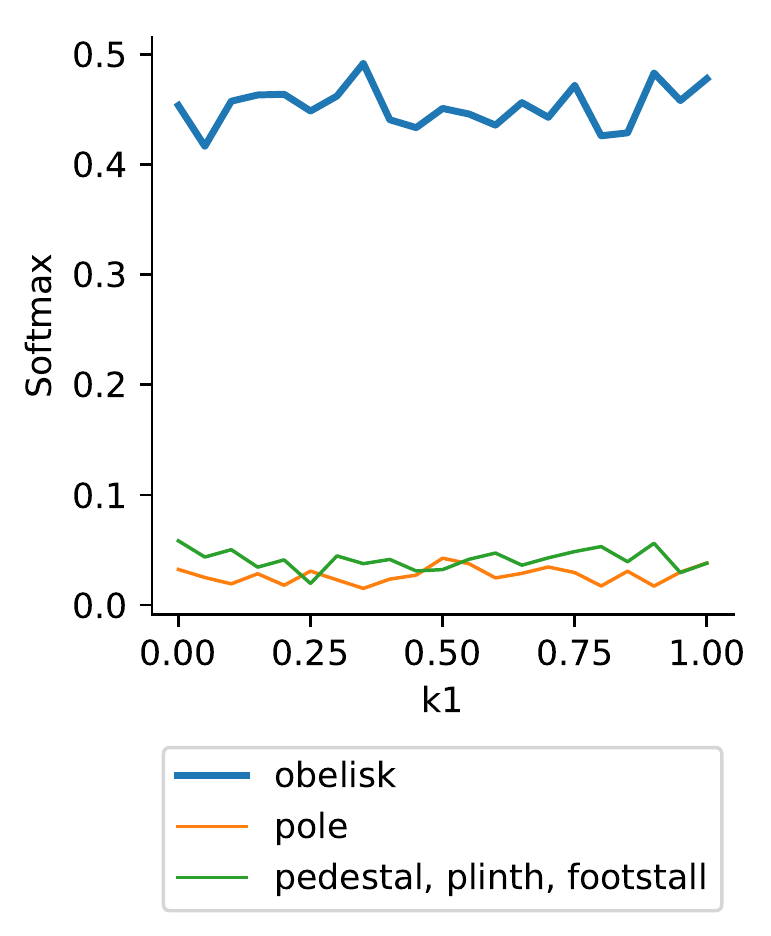}
\caption{Classification of images transformed with $F([k_1,0],[0,0])$.}
\label{fig:class_inv}
\end{subfigure}%
\caption{\textbf{Results of the training process of the \textit{Invariant Transformer Net}.}}
\label{}
\end{figure*}

In the second approach we tried to find transformations $F(k_1,k_2)$ by training the \textit{Invariant Transformer Net} with images of the validation set of \textit{ImageNet} of different classes \cite{krizhevsky2012}. The training shows a constant mean of the loss of the CNN but a reducing loss of the transformers of the \textit{Invariant Transformer Net} as we expect, since we want to find transformations of high magnitude (fig. \ref{fig:training}).

In figure \ref{fig:class_inv} we show that the classification of the transformed images does not change and thus, the network is expected to be invariant to the learned transformations.

\begin{figure*}[t]
\centering
\begin{subfigure}[t]{0.23\textwidth}
\includegraphics[width=\textwidth]{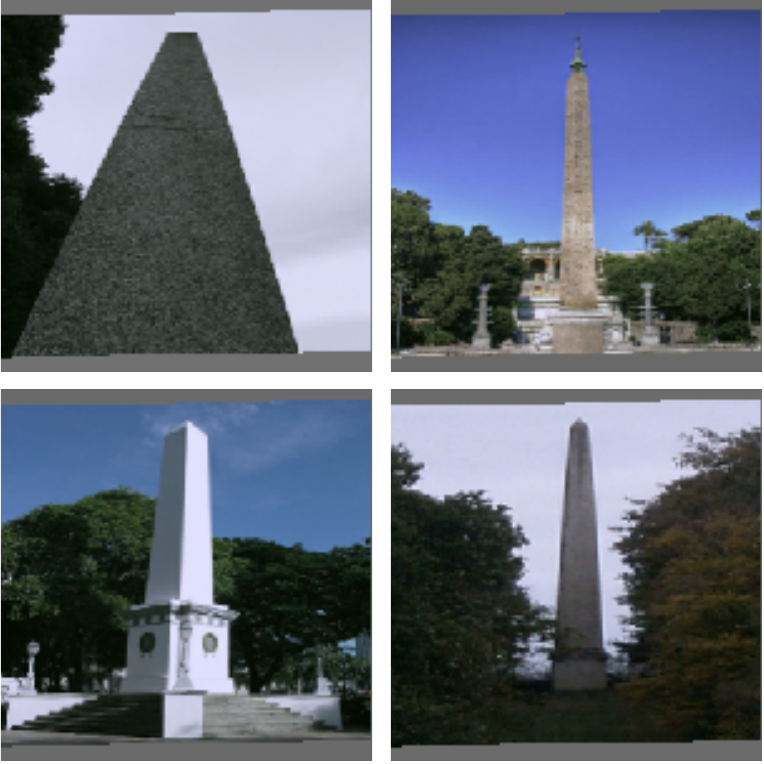}
\caption{$F([1,0], [0,0])$}
\label{fig:plots_k_1}
\end{subfigure}%
\begin{subfigure}[t]{0.23\textwidth}
\includegraphics[width=\textwidth]{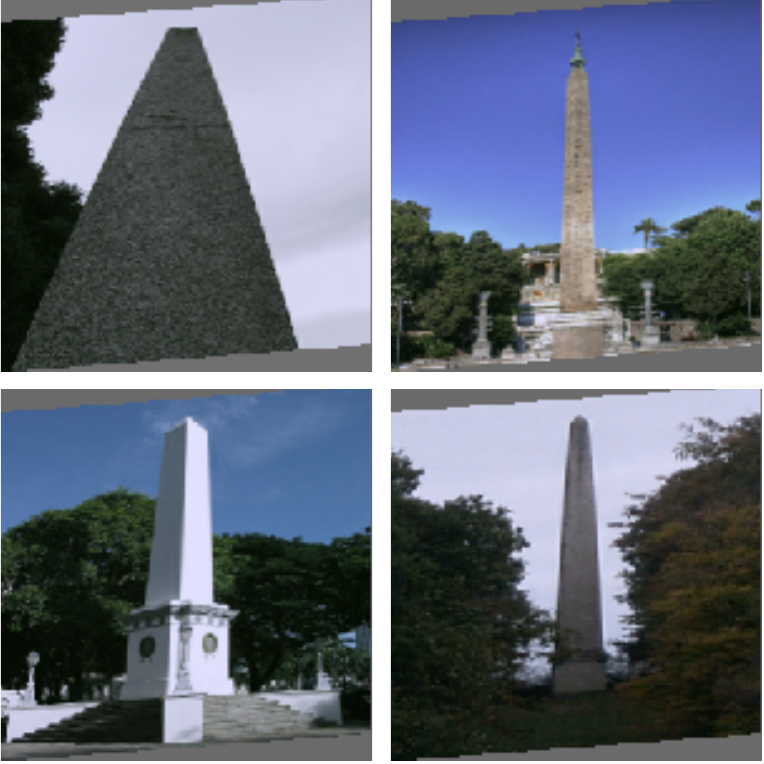}
\caption{$F([0,1], [0,0])$}
\label{fig:_k_2}
\end{subfigure}%
\begin{subfigure}[t]{0.23\textwidth}
\includegraphics[width=\textwidth]{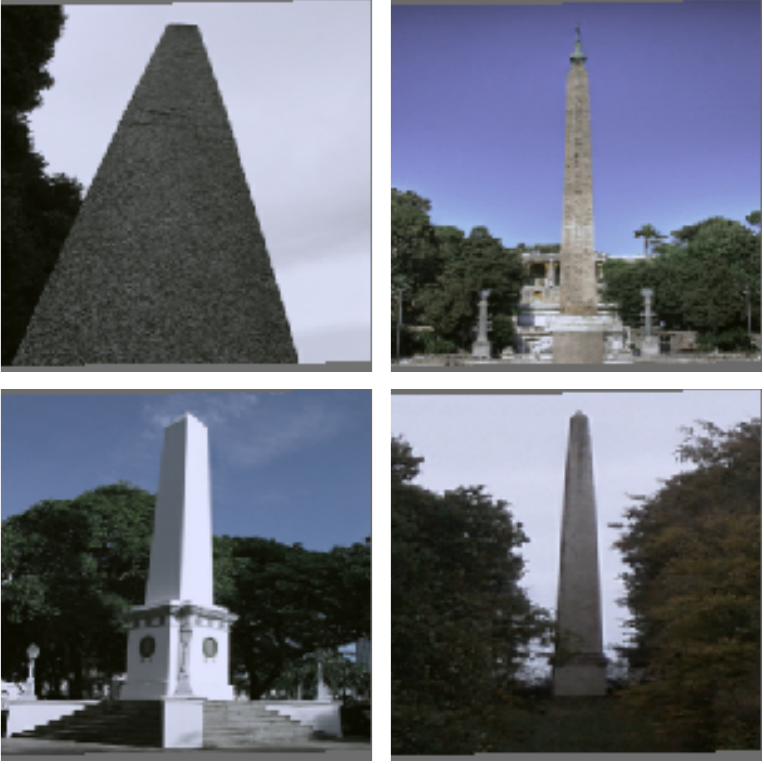}
\caption{$F([0,0], [1,0])$}
\label{fig:_k_3}
\end{subfigure}%
\begin{subfigure}[t]{0.23\textwidth}
\includegraphics[width=\textwidth]{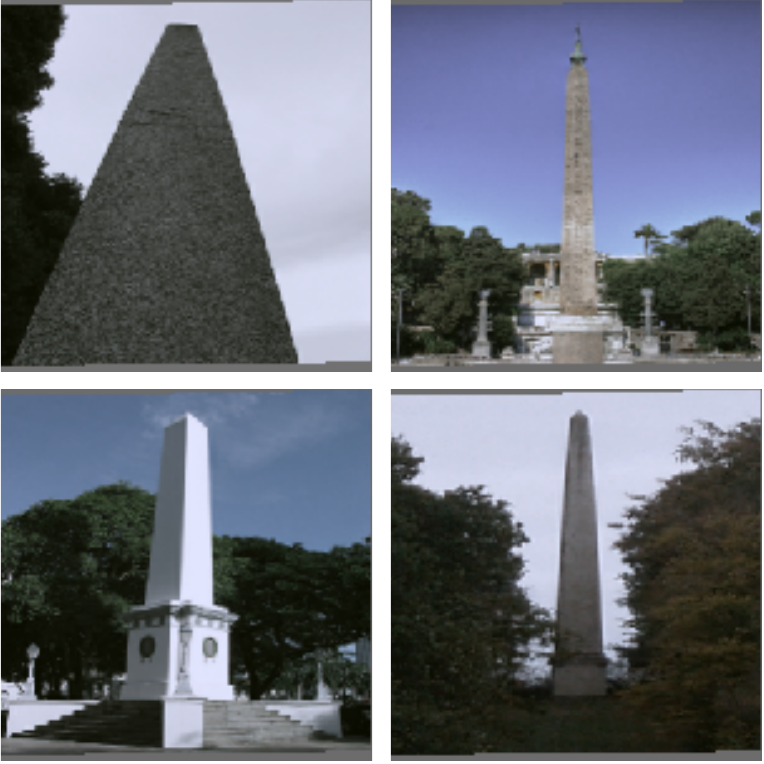}
\caption{$F([0,0], [0,1])$}
\label{fig:_k_4}
\end{subfigure}
\caption{\textbf{\textit{Invariant Transformer Net} transformations.} Each set of four images shows one parameter instance of $k_1$ and $k_2$.}
\label{fig:plots_itn}
\end{figure*}

Some of the actual transformations can be seen in figure \ref{fig:plots_itn}. Note that changes of the parameter $k_1$ result in a changing spatial transformation (with increasing magnitude for increasing $k_1$) and changes in $k_2$ result in a change of color transformation.

\section{Discussion}
The results of the large scale screening, as described in the first strategy, are consistent with the behavior one expects from common CNN architectures. This approach represents a general method to systematically access the invariances learned by CNNs and to extract thresholds at which the magnitude of different transformations lead to a misclassification of the input images. We showed that the learned invariances of \textit{ResNet} correlate to the invariance learned by \textit{AlexNet} and that both networks are highly sensible to stronger affine and nonaffine transformations.

The second strategy reveals interesting insights: While the network is able to learn small transformations of the input, it never chooses a transformation with high information loss, and thus, never strongly zooms into the image or rotates it more than a few degrees. Contrary, it only zooms out of the image which seems to only compress the image without changing  too much.
Additionally, Convolutional Neural Networks are highly sensible to color changes \cite{engilberge2017}. This might be a reason why only color changes of rather low magnitude are learned.

Differently from the approach Karel Lenc \textit{et al.} proposed in 2015, we focused on the questions to which magnitude of transformations CNNs are invariant \cite{lenc2015}.

For future work, it would be interesting to see different transformations learned with the \textit{Invariant Transformer Net} approach described above. For example, one could also learn parametrized convolutions on input images.

\section{Summary}
This paper introduced the idea of learning the space of different affine transformation families in which the modified images are still correctly classified. The architecture of the \textit{Invariant Transformer Net} can be used with different CNN architectures and allows to control the transformation via differentiable parameters, which are passed as inputs to the network.
Furthermore, the large scale screening of affine and nonaffine transformations showed the invariances architectures like \textit{AlexNet} and \textit{ResNet} learned. If the magnitude of different transformations exceeds a class- and transformation dependent threshold, the prediction result is instable and incorrect.

\printbibliography

\clearpage
\pagebreak
\onecolumngrid
\appendices
\section{Sensibility To Transformations}
Results showing effects of affine and nonaffine transformations on classification results achieved by the \textit{AlexNet} and \textit{ResNet} architectures.
\begin{figure}[H]
\centering
\begin{subfigure}[t]{0.23\textwidth}
\includegraphics[width=\textwidth]{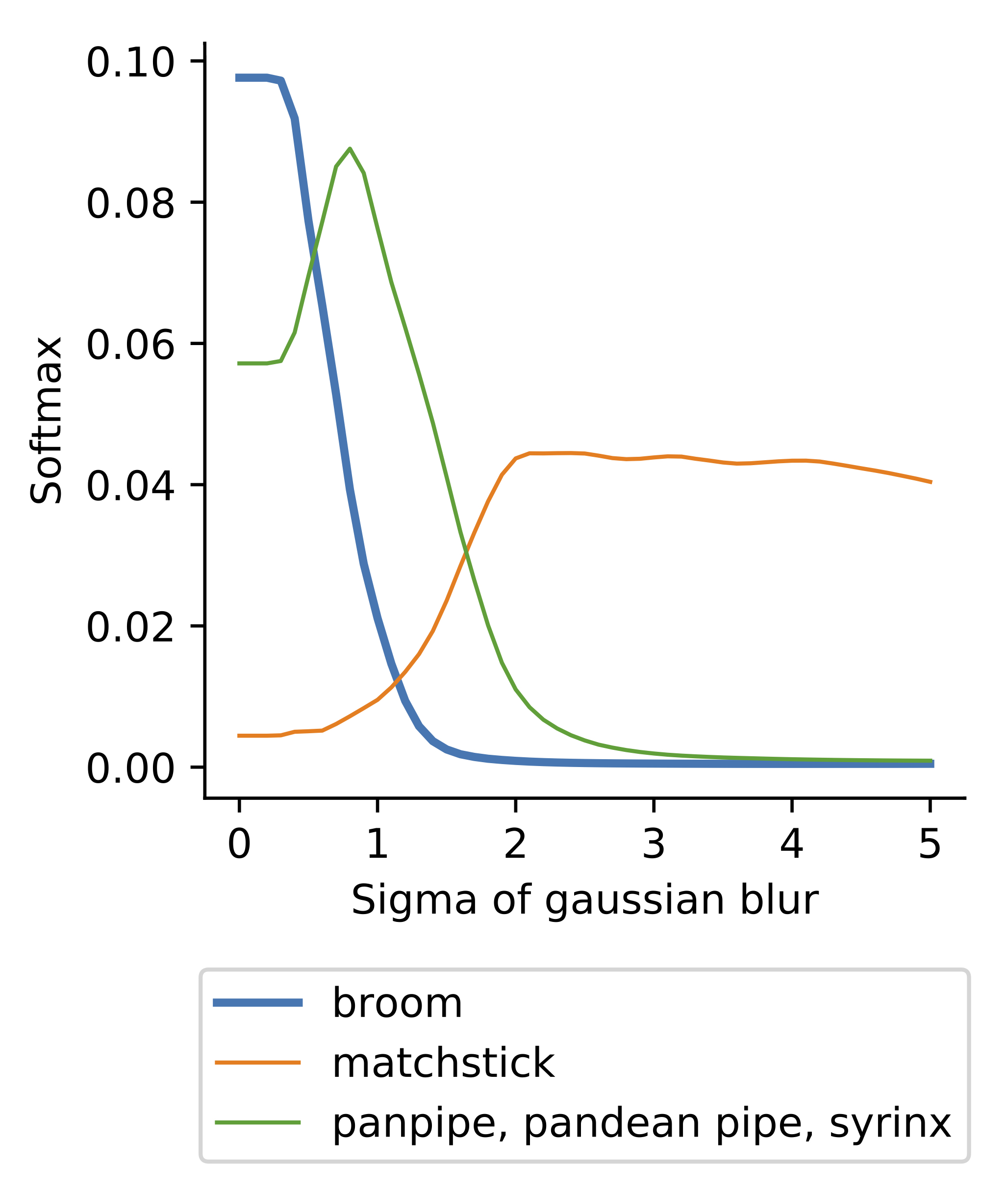}
\caption{Effect of increasing addition of Gaussian blur on the \textit{broom} class.}
\label{fig:appendix_alex_blur}
\end{subfigure}%
\begin{subfigure}[t]{0.23\textwidth}
\includegraphics[width=\textwidth]{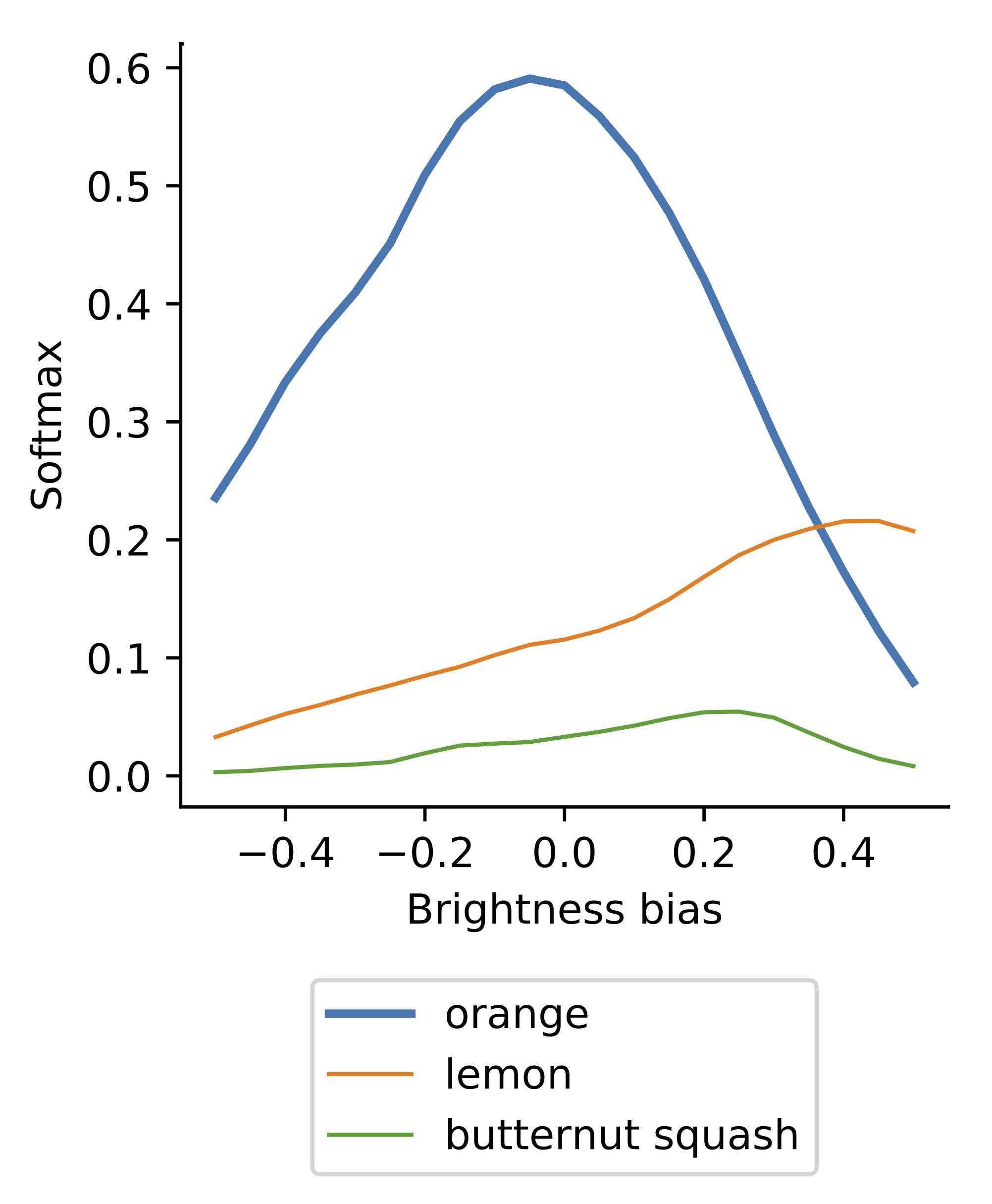}
\caption{Effect of brightness on the \textit{orange} class.}
\label{fig:appendix_alex_bright}
\end{subfigure}%
\begin{subfigure}[t]{0.23\textwidth}
\includegraphics[width=\textwidth]{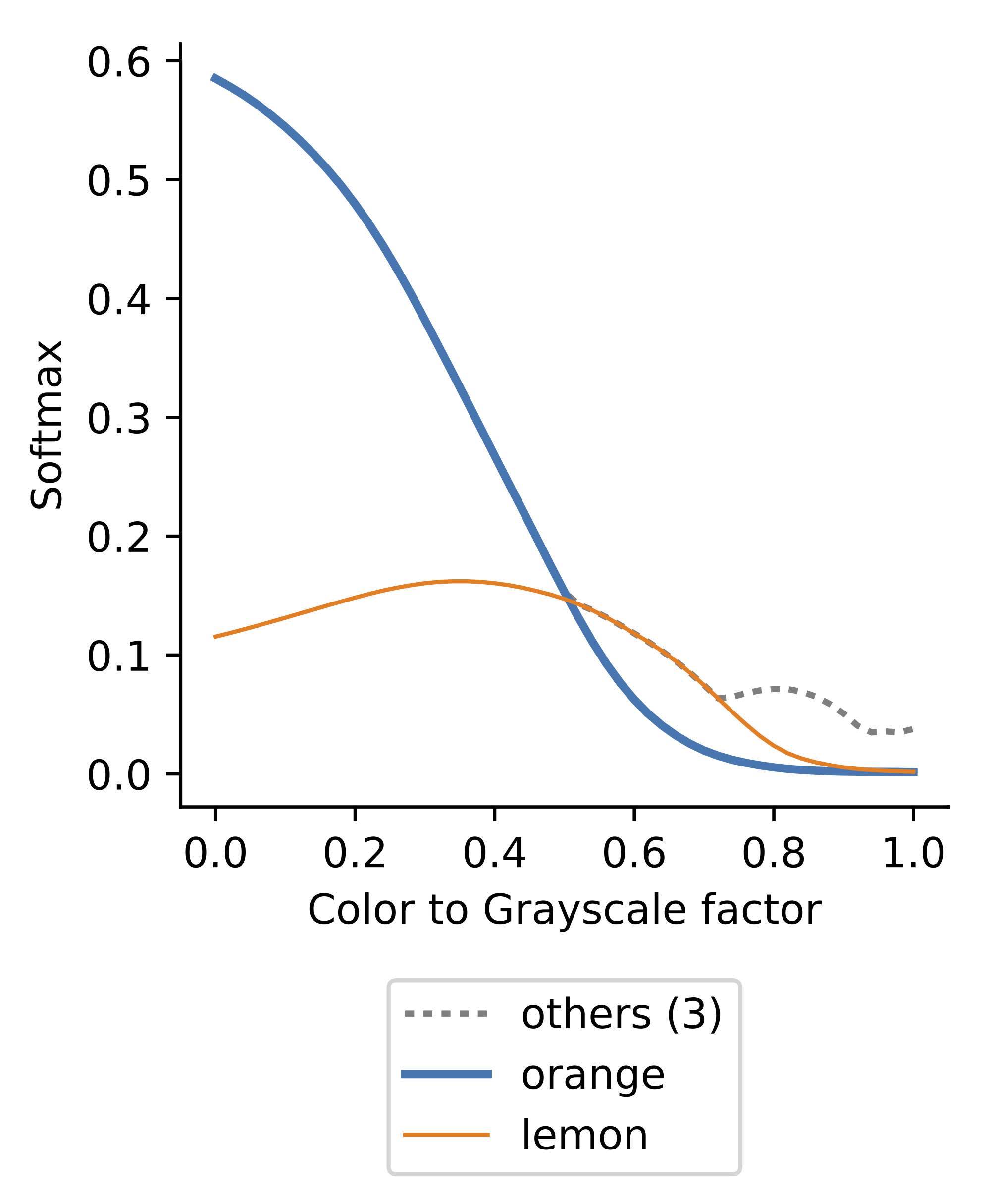}
\caption{Effect of grayscale on the \textit{orange} class.}
\label{}
\end{subfigure}%
\begin{subfigure}[t]{0.23\textwidth}
\includegraphics[width=\textwidth]{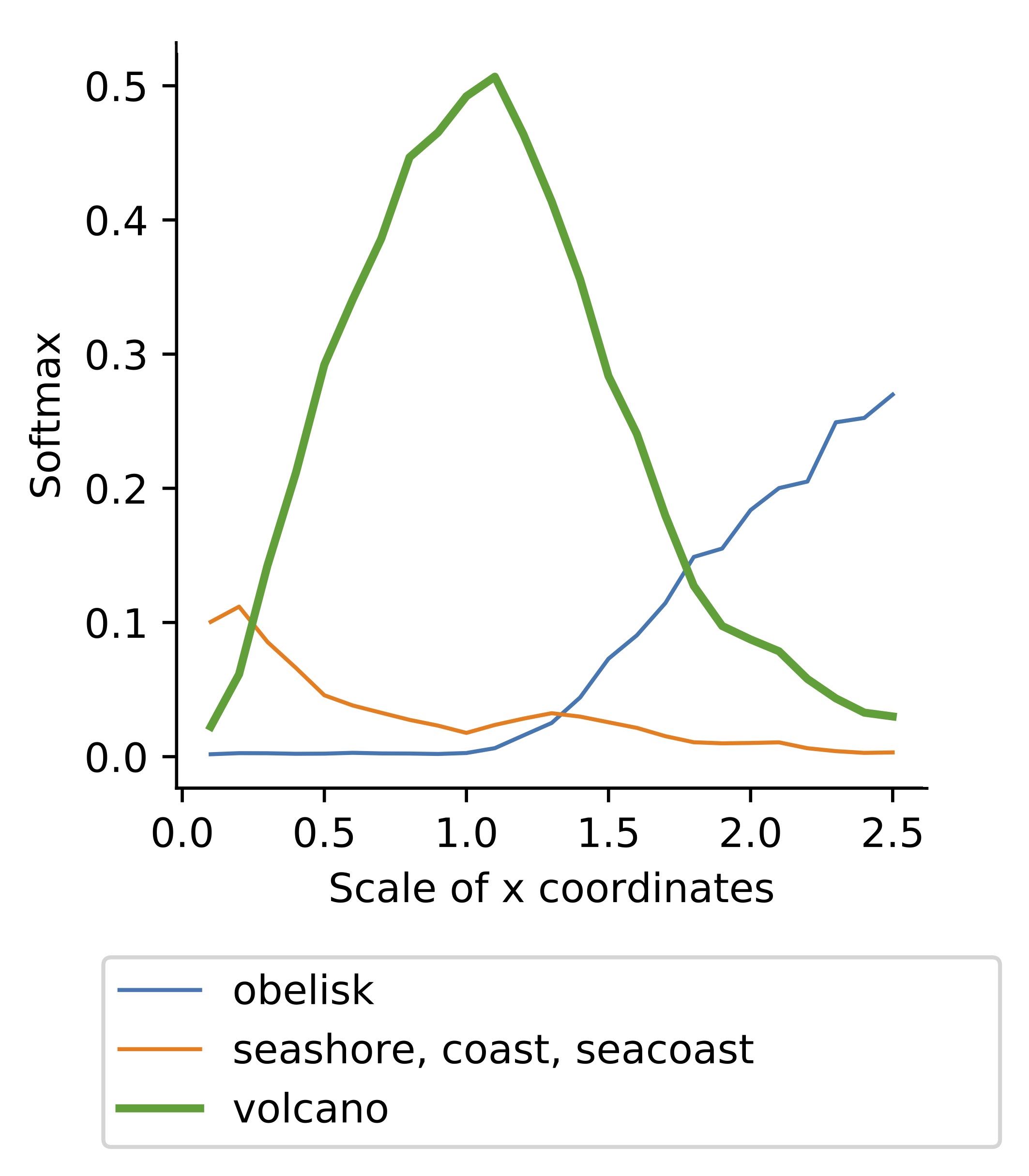}
\caption{Effect of scaling on the \textit{volcano} class.}
\label{}
\end{subfigure}
\caption{\textbf{Effect of affine and nonaffine transformations on classification results achieved by the \textit{AlexNet} architecture.} Class-wise average softmax outputs of the CNN with respect to increasing effects of different transformations. Here the effect of Gaussian noise, rotation  zooming and translation is visible.}
\label{fig:appendix_alex}
\end{figure}
\begin{figure}[H]
\centering
\begin{subfigure}[t]{0.23\textwidth}
\includegraphics[width=\textwidth]{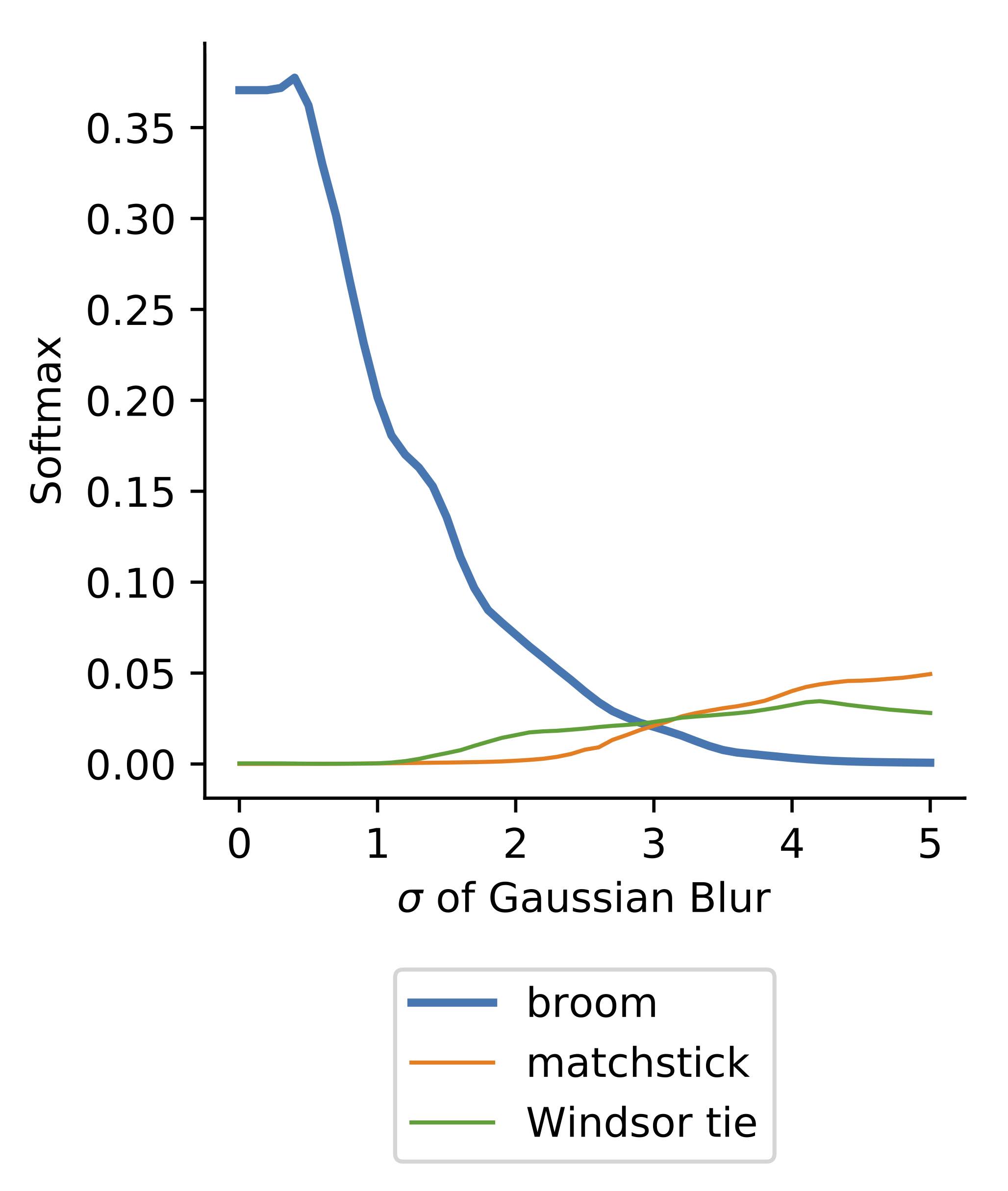}
\caption{Effect of increasing addition of Gaussian blur on the \textit{broom} class.}
\label{fig:appendix_alex_blur}
\end{subfigure}%
\begin{subfigure}[t]{0.23\textwidth}
\includegraphics[width=\textwidth]{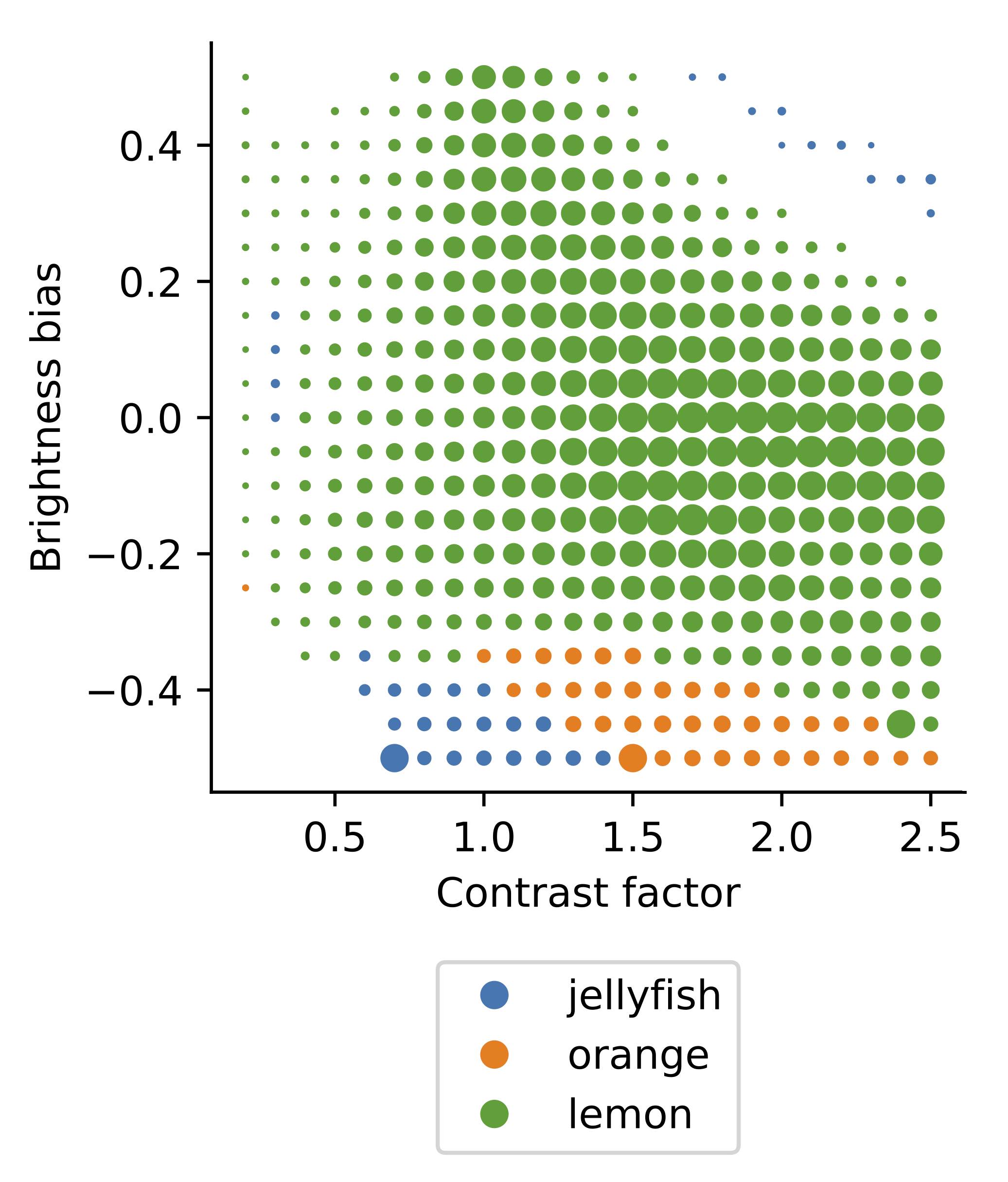}
\caption{Effect of brightness on the \textit{orange} class.}
\label{fig:appendix_alex_bright}
\end{subfigure}%
\begin{subfigure}[t]{0.23\textwidth}
\includegraphics[width=\textwidth]{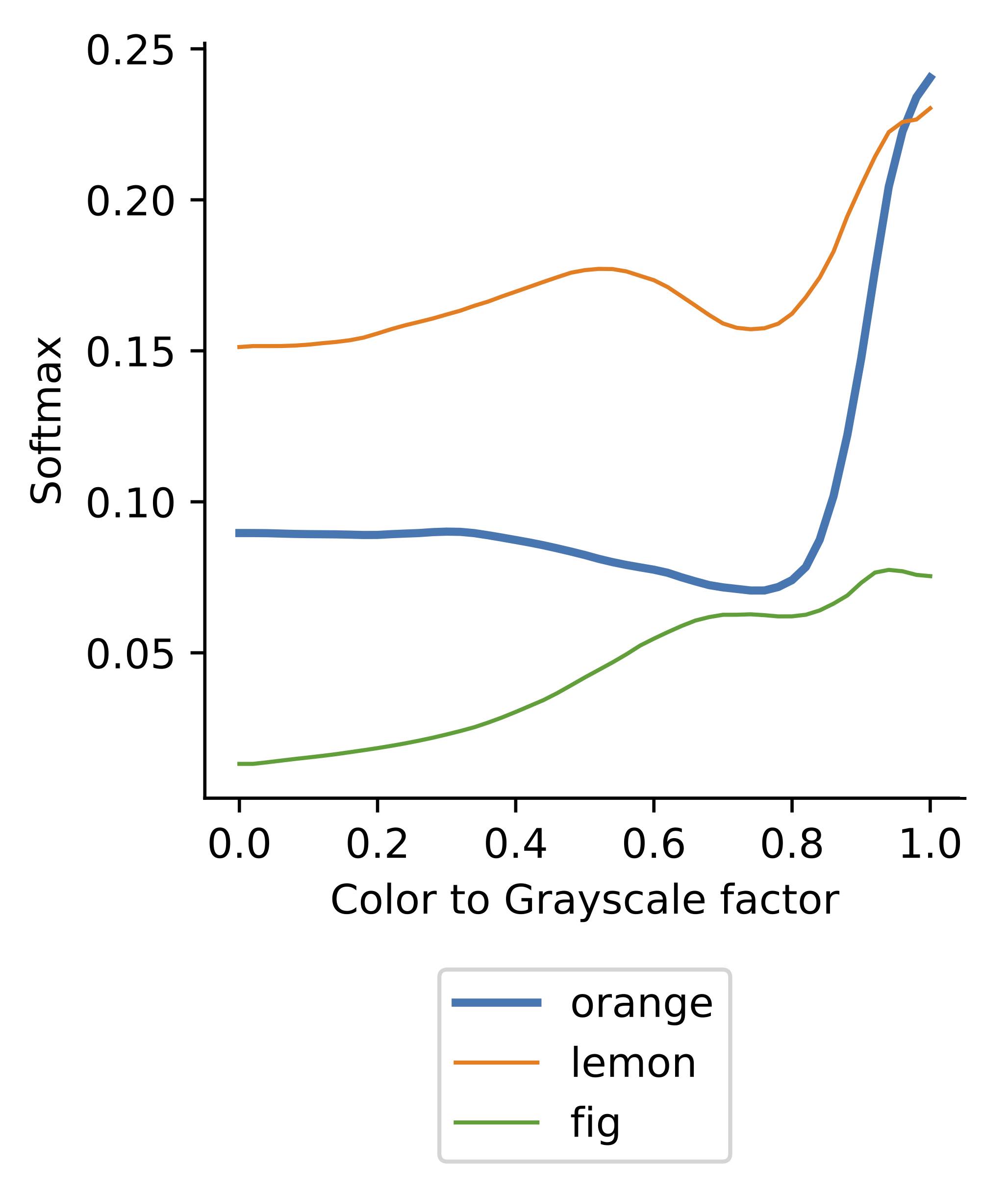}
\caption{Effect of grayscale on the \textit{orange} class.}
\label{}
\end{subfigure}%
\begin{subfigure}[t]{0.23\textwidth}
\includegraphics[width=\textwidth]{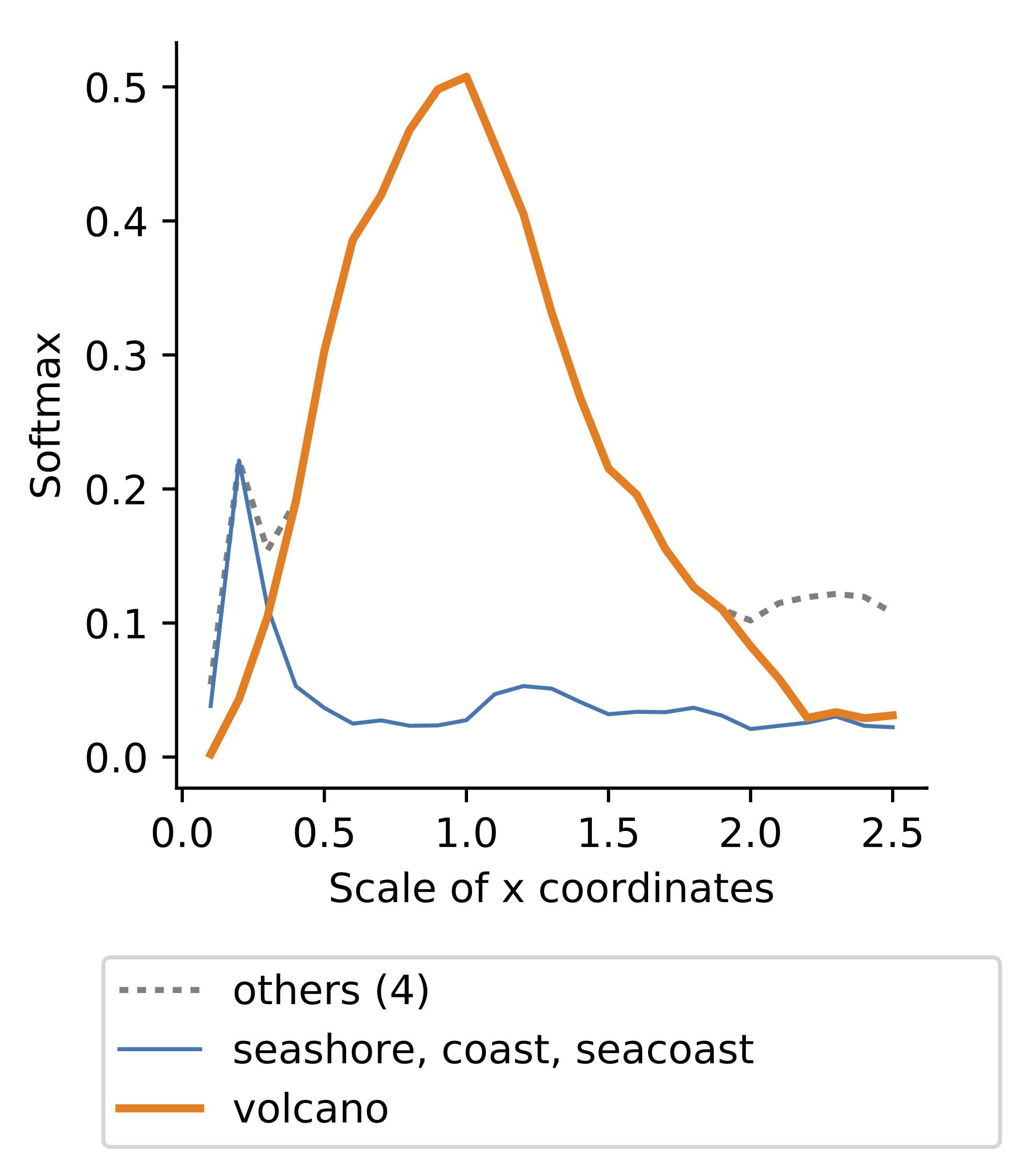}
\caption{Effect of scaling on the \textit{volcano} class.}
\label{}
\end{subfigure}
\caption{\textbf{Effect of affine and nonaffine transformations on classification results achieved by the \textit{ResNet} architecture.} Class-wise average softmax outputs of the CNN with respect to increasing effects of different transformations. Here the effect of Gaussian noise, rotation  zooming and translation is visible.}
\label{fig:appendix_alex}
\end{figure}
\clearpage
\end{document}